\documentclass[letterpaper]{article}
\usepackage{natbib,alifeconf}
\newcommand{\ul}{\underline}
\usepackage{multicol}
\usepackage{graphicx}

\usepackage{hyperref}

\title{Markov Brains: A Technical Introduction}
\author{Arend Hintze$^{1,2,3}$, Jeffrey A. Edlund$^{4}$, Randal S. Olson$^5$, David B. Knoester$^{3,10}$,\\
Jory Schossau$^{1,2,3}$, Larissa Albantakis$^6$, Ali Tehrani-Saleh$^{1,3}$\\
Peter Kvam$^{9}$, Leigh Sheneman$^{1,3}$, Heather Goldsby$^{1,2,3}$, Clifford Bohm$^{2,3}$ \and Christoph Adami$^{3,10,11}$ \\
\mbox{}\\
$^1$Department of Computer Science \& Engineering, Michigan State University\\
$^2$Department of Integrative Biology, Michigan State University \\
$^3$BEACON Center for the Study of Evolution in Action, Michigan State University\\
$^4$Computation \& Neural Systems, California Institute of Technology\\
$^5$Institute for Biomedical Informatics, University of Pennsylvania\\
$^6$Department of Psychiatry, University of Wisconsin\\
$^9$Department of Psychological \& Brain Sciences, Indiana University \\
$^{10}$Department of Microbiology \& Molecular Genetics, Michigan State University\\
$^{11}$Department of Physics \& Astronomy, Michigan State University\\
}

\begin{document}
\maketitle

\begin{abstract}
Markov Brains are a class of evolvable artificial neural networks (ANN). They differ from conventional ANNs in many aspects, but the key difference is that instead of a layered architecture, with each node performing the same function, Markov Brains are networks built from individual computational components. These computational components interact with each other, receive inputs from sensors, and control motor outputs. The function of the computational components, their connections to each other, as well as connections to sensors and motors are all subject to evolutionary optimization. Here we describe in detail how a Markov Brain works, what techniques can be used to study them, and how they can be evolved. 
\end{abstract}

\section{Introduction}
\begin{figure}[!htb]
\begin{center}
\includegraphics[width=3in]{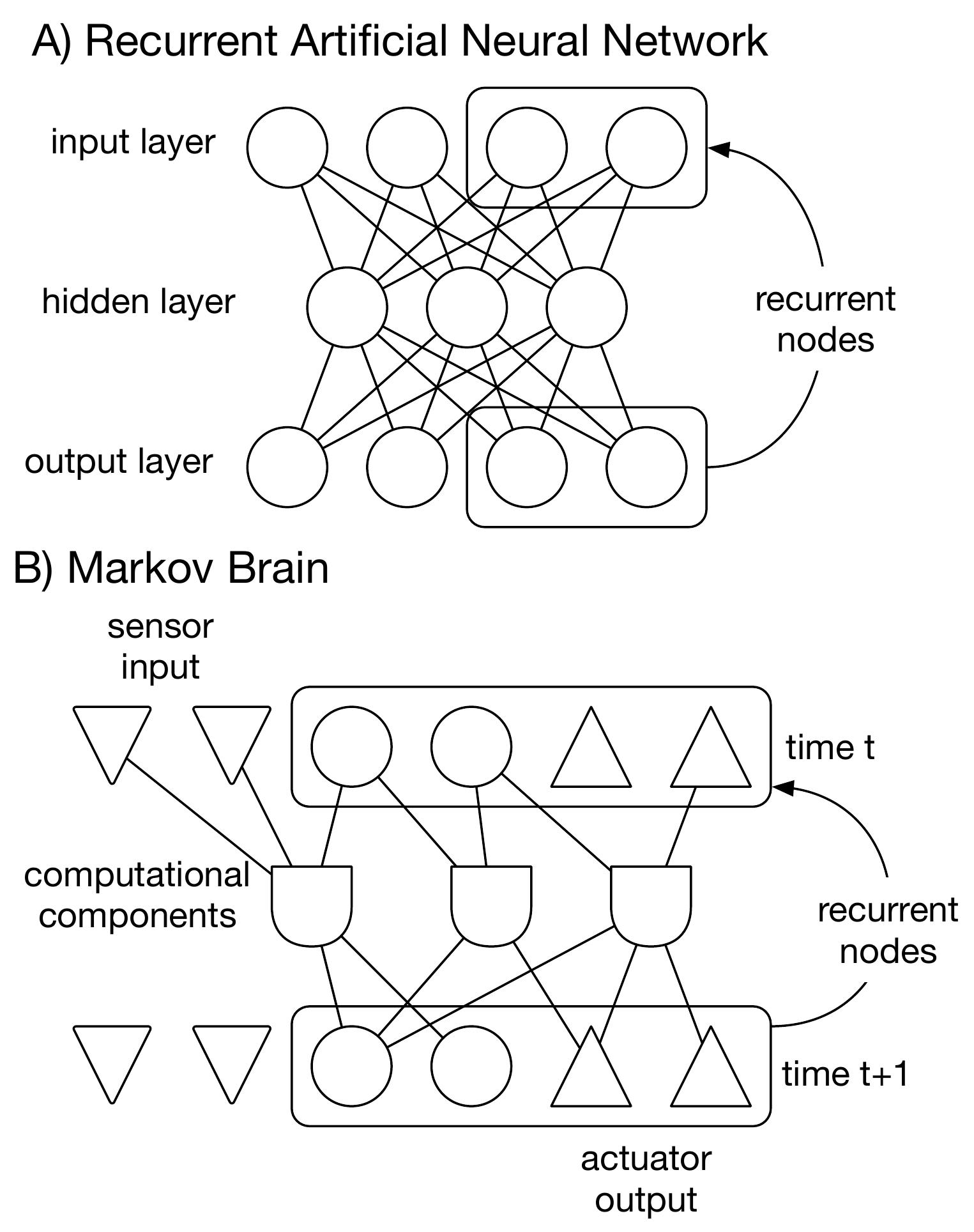}
\caption{Illustration of a recurrent artificial neural network and a Markov Brain. A) The recurrent artificial neural network has an input layer that contains additional recurrent nodes. All nodes of each layer are connected to all nodes of the next layer. ANNs can have hidden layers, but that is not a requirement. Connections between nodes typically specify a connection weight, in most cases all nodes of an ANN are updated using the same transfer and threshold function, but that is also not a requirement. A special class of ANNs that do not have this topology are NEAT and HyperNEAT ANNs \citep{gauci2010autonomous} which also have an evolvable topology, as well as an evolvable transfer and threshold function. B) A MB is defined by a state buffer (sometimes called state vector) where each element represents one node. Inputs are written into a part of that buffer. A Brain update applies a set of computations defined by the {\em computational components} (sometimes referred to as ``logic gates") on this buffer. Outputs are read from a subset of the buffer. Since inputs are generally overridden by the environment they should not be considered part of what is recurring. However, the outputs of the last update are generally accessible by the computational components and are thus part of the recurrence. Here computational components are illustrated using the same shape, however MBs 
in practice are evolved using different types of computational components.}
\label{fig:ANNvsMB}
\end{center}
\end{figure}

Artificial neural networks~\citep{russell1995modern} (ANNs) have become a staple for classification problems as well as a common tool in neuro-evolution research. Here we assume that the reader knows how a simple three layer perceptron with input, hidden, and output layers works, what a recurrent neural network is, how an ANN can be used to control a robot or an embodied virtual agent, and what a perception action loop is (see Figure \ref{fig:ANNvsMB}). As we will discuss methods to evolve Markov Brains (MB)\footnote{In this report we use a capital B for brains that are specifically artificial brains of the Markov type. Biological brains continue to be just brains.}, we further assume that the reader has general knowledge of genetic algorithms and digital evolution. 

The structure and function of a MB can be encoded by a genome, where mutations lead to functional and topological changes of the network. In this case, evolution occurs on the level of the genotype. It is also possible to encode MBs directly, which means that the network itself evolves by experiencing rewiring and functional changes as well as additions and subtractions of computational components, governed by specific rules. We therefore distinguish between \textit{genetically} and \textit{directly encoded} MBs. 
Regardless of the encoding, a MB always is defined by a state buffer that contains all input, hidden, and output states. This buffer is sometimes referred to as the state vector, and we call individual components of this vector ``states", or ``nodes". This buffer can experience an update where the computational components read from this buffer and compute the next state of the buffer (called $t\to t+1$ update). Each of these Brain updates happen in parallel (even though one can also perform sequential computations if so desired). An agent is controlled by computing what the agent senses, feeding this data into the input states, performing a Brain update, reading from the output states of the buffer, and using these outputs to control the agent in the environment. This constitutes one whole perception-action loop, and if repeated, allows the MB to control an agent over a defined period or lifetime. There is no explicit restriction or definition on the number of Brain or world updates to be computed within one whole action-perception loop. For example, it is possible to update the world once, followed by ten Brain updates. The experimenter needs to decide this, and different applications might require different settings.

The logic components do not necessarily read from all states of the buffer (like one layer of an ANN always reads from the previous layer),  instead each component has a subset of connections that can select a subset of states from the buffer. In addition, each logic gate can write into many different states of the buffer. The connections can change over the course of evolution, and in theory one can conceive of a gate that changes its connectivity even during the lifetime of a MB, for example, in an implementation that features learning.

Connecting the outputs of computational components arbitrarily to the hidden states creates situations in which a single state of the buffer becomes updated by more than one computational component. Computational components being executed sequentially would lead to states being overridden. However, computational components are updated in parallel, and thus states that receive multiple inputs integrate their signal by summation. (In older implementations the states were binary and then an OR function was applied instead of the summation used now). We think of the computational components as roughly analogous to neurons, and the zeros and ones of the binary logic operations as the absence or presence of an action potential. Along an axon you either do, or do not, have an action potential, and thus we think that a summation resembles this phenomenon best. 

Computational components are allowed to write into input states of the buffer (as well as read from output states) it just so happens that those input states will be overridden by the world once a new percept is computed. Reading from output states (proprioception) is sometimes not desired and then output states are set to $0$ before a new Brain update is computed.

The state buffer is a vector of continuous variables (64 bit double precision). At the same time, many earlier implementations used only discrete binary values as states ($0$ and $1$). Therefore, gates that require binary inputs use a discretizer that interprets all values larger than $0$ as a $1$, while all other values become a $0$. 

Any type of computational component needs two types of values to be defined properly. All the values that define function and connectivity as well as the ranges within these values can vary. While evolution optimizes the values that define a computational component the ranges are set by the experimenter (which we call parameters). For example, the experimenter could limit the number of input connections to $2$ and the number of output connections to $8$. If a mutation to the number of inputs or outputs occurs, the new value will be kept within those predefined conditions. How these values are encoded by a genome can vary between implementations, and also varies between the different kinds of components. 

\subsection{Computational Components}
MBs are made from different types of computational components that interact with each other and the outside world. Many different such components are available, and new such units are being designed that will be added to this document on a regular basis. Earlier publications using MB technology used fairly simple components, but there is technically no limit to what these components can be. Traditionally, MBs used deterministic or probabilistic logic gates as their computational building blocks. We added components that simulate classical artificial neural networks, as well as thresholding functions, timers and counters. We defined logic gates that can process feedback and perform lifetime learning, as well as gates that resemble biological neurons and gates that can perform ternary logic operations. Here we present each of these components. Gates can read from multiple inputs and write into multiple outputs -- how this extends the logic of a conventional logic gate is explained in the section for deterministic logic gates, but applies to all other gates if not stated otherwise.

\subsection{Deterministic Logic}
Logic gates are commonly known to perform the essential functions of computation, such as: AND, NOT, OR, XOR, NAND, etc. These gates typically receive two binary inputs (or one in the case of the NOT gate) and return the result of their computation as a single bit. This constitutes what we call a 2-in-1-out gate. However, logic gates can be built for more inputs and outputs. A multi-AND might have eight inputs and only returns a $1$ (true) if all its eight inputs are also $1$ (true) and a $0$ (false) in all other 255 possible input cases. This would be an example of a 8-in-1-out gate. A three-bit multiplexer has three inputs and eight outputs (3-in-8-out gate) and the three inputs control which of the eight output wires is on. Similarly, the deterministic logic gates that MBs use can have an arbitrary number of inputs and outputs. We typically limit them to four inputs and four outputs maximally, and require at least one input and one output.

The logic of the gate is defined by a logic table. An AND gate with the inputs $I_{A}$ and $I_{B}$ and a single output $O$, for example, would have the following logic table:

\begin{table}[!hbt]
\center{
\begin{tabular}{|c|c||c|}\hline
$I_{A}$ & $I_{B}$ & $O$\\ \hline
0 & 0 & 0\\
0 & 1 & 0\\
1 & 0 & 0\\
1 & 1 & 1\\ \hline
\end{tabular}
}
\vskip 0.25cm
\caption{A 2-in-1-out logic table for an AND gate}
\end{table}
This becomes a more complicated table if multiple outputs are involved. Imagine a 2-in-2-out deterministic logic gate with the two inputs $I_{A}$ and $I_{B}$, as well as the two outputs $O_{A}$ and $O_{B}$. The resulting logic table would become:

\begin{table}[!hbt]
\center{
\begin{tabular}{|c|c||c|c|}\hline
$I_{A}$ & $I_{B}$ & $O_{A}$ & $O_{B}$\\ \hline
0 & 0 & 0 & 0\\
0 & 1 & 0 & 1\\
1 & 0 & 0 & 1\\
1 & 1 & 1 & 0\\ \hline
\end{tabular}
}
\vskip 0.25cm
\caption{Logic table for a 2-in-2-out gate, where output $O_{A}$ is following the AND logic for the inputs, while output $O_{B}$ is the XOR logic}
\end{table}
However, as we will later need to define probabilistic logic tables (see below), we use a different notation that defines both outputs at the same time. An example of such a table for the same 2-in-2-out gate as above is:
\begin{table}[!hbt]
\center{
\begin{tabular}{|c|c||c|c|c|c|}\hline
$I_{A}$ & $I_{B}$ & 0,0 & 0,1 & 1,0 & 1,1 \\ \hline
0 & 0 & 1 & 0 & 0 & 0\\
0 & 1 & 0 & 1 & 0 & 0\\
1 & 0 & 0 & 1 & 0 & 0\\
1 & 1 & 0 & 0 & 1 & 0\\ \hline
\end{tabular}
}
\vskip 0.25cm
\caption{Logic table for a 2-in-2-out gate, where output $O_{A}$ is following the AND logic for the inputs and output $O_{B}$ has the XOR logic. The output of $O_{A}$ and $O_{B}$ is now noted in the same column. For example if both inputs are $0$ both outputs must be $0$ as well, indicated by the column labeled 0,0 containing a 1 for and input of 0,0. Similarly, if both inputs are 1, output $O_{A}$ must be 1 due to the AND logic, while output $O_{B}$ must be 0 due to the XOR logic. This is indicated as output 1,0, as seen in the column title}
\end{table}
The same principle applies to larger logic gates.

\subsection{Probabilistic Logic Gates}
Probabilistic logic gates work similarly to deterministic logic gates in that they compute a mapping between a set of inputs and outputs. However, this mapping is probabilistic. Instead of using a deterministic logic table, we use a probabilistic one, which means that we can use the same table with deterministic mappings (1s and 0s) swapped out for stochastic ones ($p$ ranging from 0.0 to 1.0). In this case, the output table is interpreted as a matrix of probabilities. The probability matrix for the 2-in-2-out table from above defining an AND and an XOR would look like this:

\begin{table}[!hbt]
\center{
\begin{tabular}{|c|c||c|c|c|c|}\hline
$I_{A}$ & $I_{B}$ & $0,0$ & $0,1$ & $1,0$ & $1,1$ \\ \hline
0 & 0 & 1.0 & 0.0 & 0.0 & 0.0\\
0 & 1 & 0.0 & 1.0 & 0.0 & 0.0\\
1 & 0 & 0.0 & 1.0 & 0.0 & 0.0\\
1 & 1 & 0.0 & 0.0 & 1.0 & 0.0\\ \hline
\end{tabular}
}
\vskip 0.25cm
\caption{Probabilistic logic table for a 2-in-2-out gate, where output $O_{A}$ follows AND logic for the inputs and output $O_{B}$ follows XOR logic. Note that all probabilities are 0.0 and 1.0 only and thus, even though we use probabilities, the logic remains deterministic. Each column represents the probability for one of the four possible outputs to occur.}
\end{table}

We can see that deterministic logic gates are a special case of probabilistic logic gates. At the same time, we can create probability matrices that are not deterministic, like the one shown in the following table:

\begin{table}[!hbt]
\center{
\begin{tabular}{|c|c||c|c|c|c|}\hline
$I_{A}$ & $I_{B}$ & $P(0,0)$ & $P(0,1)$ & $P(1,0)$ & $P(1,1)$ \\ \hline
0 & 0 & 0.1 & 0.2 & 0.3 & 0.4\\
0 & 1 & 0.5 & 0.1 & 0.2 & 0.2\\
1 & 0 & 0.1 & 0.1 & 0.1 & 0.7\\
1 & 1 & 0.25 & 0.25 & 0.25 & 0.25\\ \hline
\end{tabular}
}
\vskip 0.25cm
\caption{Probabilistic logic table for a 2-in-2-out gate, where output $O_{A}$ is following the AND logic for the inputs, while output $O_{B}$ is the XOR logic.}
\end{table}
Let us use introduce the mathematical definitions here: Each gate has a defined number of inputs ($n_{I}$) and a defined number of outputs ($n_{O}$). We call the input a gate receives $I$, where $I$ can range from $0$ to $2^{n_{O}}$ since inputs are binary and this range defines all possible combinations. Similarly, there are at most $2^{n_{O}}$ possible outputs which we call $O$. At every update a gate receives an input $I$ and needs to produce an output  $O$ defined by a probability matrix we call $P$. This table has the dimension of $2^{n_{I}} * 2^{n_{O}}$, where each element of this matrix $P_{IO}$ defines the likelihood of an input $I$ resulting in an output $O$. As a consequence, each row of the matrix $P$ needs to sum to $1.0$:

\begin{equation}
    1.0=\sum_{j=0}^{j<2^{n_{O}}} P_{Ij}
\end{equation}
When this matrix is read from a genome, for example, the table will be filled with arbitrary values read from the genome that will automatically be normalized. If ever a row would be filled with $0.0$ everywhere, the row would become a row of equal probabilities.

\subsection{ANN Gate}
This type of gate also has a flexible number of inputs and outputs. It implements a single-layer artificial neural network that straddles between the inputs and outputs of the gate. Each output node is connected to each input node of the gate, and each connection has a weight. The transfer function for each node $j$ is the sum over all weight input products, where $I_{max}$ defines the number of all inputs:

\begin{equation}
T_{j}=\sum_{i=0}^{i<I_{max}}I_{i}W_{ij}
\end{equation}
The standard hyperbolic tangent is used as the threshold function. 

\subsection{Threshold Gate}
The threshold gate discretizes the inputs into $0$ and $1$ and accumulates them. If their accumulated values reach a genetically determined threshold, all outputs return a $1$, and the internal accumulator is reset to zero. Otherwise this gate returns $0$ on all outputs. Note that this gate itself contains a hidden value or state.

\subsection{Timer Gate}
This gate does not need inputs, and fires a $1$ on all its outputs after a predefined number of updates passed.

\subsection{Feedback Gates}
In general feedback gates are very similar to probabilistic gates regarding inputs and outputs, as well as their probability table. However, they have two additional inputs, which are used to detect positive or negative feedback.

At any given update, this gate receives a specific input and returns a specific output depending on the evaluation of the probability table. The resulting outputs typically directly control an actuator or result in changes to the state buffer, which over time affect the actuators and thus the behavior of the agent. In turn, the actions of the agent might be beneficial or detrimental. This information is not directly given to the agent; instead, the agent needs to also evolve machinery that assesses the sensory input as positive or negative feedback. This means that at the time point the output for a gate is computed, it is not clear if the output is a good or bad one. Therefore, this gate keeps track of previous input-to-output mappings in order to apply feedback at later stages. The number of previous mappings that are remembered is genetically encoded, and thus evolvable. Specifically, for an input $I$ received at time point $t$ an output $O$ is generated for time point $t+1$. Since this gate is technically similar to the probabilistic logic gate, we also have a probability matrix $P$ (see above). The input output pair $IO$ was generated by the probability matrix $P$, specifically by the element $P_{I_{t}O_{t+1}}$. Technically, it is the entire row $P_{I_{t}}$ that is responsible, but for now we define that $P_{I_{t}O_{t+1}}$ was ''responsible``. This is very similar to the process in Q-learning with delayed rewards where this probability would be defined as $Q(s_{t},a_{t})$\citep{russell1995modern}. 

If the computation the gate performed resulted in positive feedback the probability $P_{I_{t}O_{t+1}}$ should be increased, whereas in the case of negative feedback it should be decreased. However, relevant or pertinent feedback might arrive only much later than at the time point $t+1$. On the one hand, information takes time to propagate through the MB, and on the other hand, the action of the agent might have delayed rewards. Therefore, the feedback gates store past input-output pairs $[(I_{t},O_{t+1}),(I_{t-1},O_{t+1-1}),...,(I_{t-n},O_{t+1-n})]$. The range of this buffer $n$ is an evolvable property of the gate. In case positive feedback is received, the gate sequentially increases each entry in the probability matrix defined by the $(I,O)$ pairs in the buffer. It increases these values using a random number drawn from a uniform random number between $0.0$ and a maximum value which again is evolvable. After each increase of $P_{I_{t},O_{t}}$ the row $P_{I_{t}}$ becomes normalized again, before the next change is applied.

The negative feedback mechanism works similarly, except that the value defined by $P_{I_{t},O_{t}}$ are decreased by the random value instead of increased. We prevent values from dropping below a lower threshold (typically $0.01$, which can be set by the experimenter).



\subsection{Ternary Logic Gates}
Ternary logic gates first of all discretize inputs differently. Values that are $1.0$ and above are considered to be a $1$, values $-1.0$ and lower are interpreted as a $-1$, and all values in between are interpreted as a $0$. After that, these gates function like deterministic logic gates, except that they can compute ternary logic, and can return three different values $-1$, $0$, and $1$ as their outputs. As a consequence, the logic table for a probabilistic 2$\to$1 gate, for example, is determined by 27 probabilities (three probabilities for each of the nine possible inputs).
There are different (hypothetical) reasons for using (or testing) ternary logic gates in MBs. For example, voltage-gated ion channels have three states: closed, open, and blocked~\citep{Albertsetal2002}, which implies that neurons cannot immediately fire after having fired. Furthermore, there is good evidence that information is sometimes stored in ternary logic using three different firing rates, such as in the Reichardt detectors that sense motion in the visual system (to encode motion towards the detector, motion away from the detector, and no motion). 

\section{Encoding}
So far we listed the possible components a MB can have, and one can easily imagine more. How these components are encoded genetically is crucial, as the manner of encoding will determine how the component reacts to mutations. In principle we could use a direct encoding in which mutations are applied to the Brain itself. For example, we might randomly select probabilities in the gate table to change, gates to rewire randomly, gates to disappear, or new gates to be added. But it is well-known that direct encodings are usually fragile, and all prior implementations of MBs have used a genetic encoding. 

This encoding varies slightly from implementation to implementation and we did not experience any significant change to evolvability for example. To encode a MB, we use a vector of (currently) up to 20,000 elements that is interpreted from the beginning to the end (left to right). This is analogous to a strand of DNA where each site codes for a particular nucleotide. Here, instead of nucleotides, we use numbers (in older implementation these were bytes, presently these loci can be arbitrary values whose range can be set by the experimenter). While parsing this vector, pairs of numbers are used to identify sections that encode a single gate. This idea is inspired by how the beginning of genes are marked in DNA, namely by {\em start codons} that mark the point of transcription initiation. Theoretically, it is possible to add {\em transcriptional regulation} to the expression of MB genes that either activates or deactivates genes/gates based on environmental signals, but this idea has not yet been tested. 

\begin{figure*}[!htb]
\begin{center}
\includegraphics[width=6in]{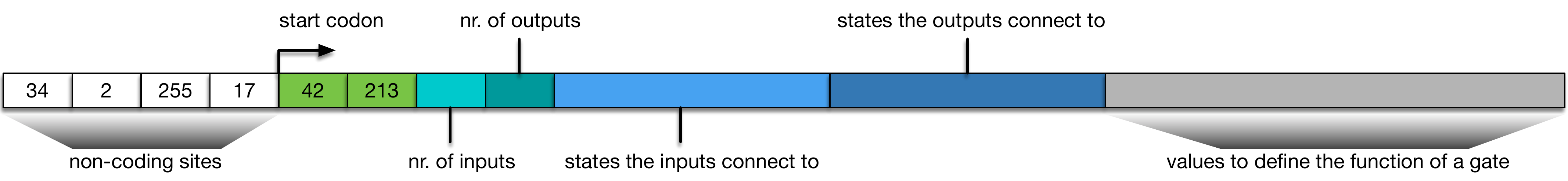}
\caption{Encoding scheme illustration. The sequence of sites of a genome is read from beginning to end. Most sites are non-coding, but when a start codon in found the following stretch of sites is defining the connections and function of a gate (analogous to a gene). After the start codon, the next sites are encoding the number of inputs and outputs, followed by sites which encode the specific nodes in the state buffer the gate receives inputs from and in which the outputs of the gate write. The last stretch of sites is used to encode the specific values needs to specify the function of each gate.}
\label{fig:EncodingExample}
\end{center}
\end{figure*}

Once a start codon is found in the genome, the stretch of sites following it is used to define all values relevant for the gate. This stretch is called a gene to stay within the biological metaphor. Since there is a wide variety of gates, how each gene is decoded is subject to the gate it defines. Generally the first pair of sites defines the number of inputs and outputs the gate has. Since the minimal and maximal number of connections for the inputs and outputs can be defined by the experimenter, the value that is read from each site it transformed into a value between the allowed minimum and maximum. The next sections define the nodes of the state buffer to be read from and written into. After that, the diversity in gates becomes so large that there is no common denominator except that the following stretch is responsible to encode the function of each gate. In the case of a probabilistic logic gate it would define the probability table, whereas a timer gate interprets the next value as the interval in which it should signal. This also implies that different gates can have different sizes on the genome. For a general description how a gene encodes a gate see Figure \ref{fig:EncodingExample}.

The start codon used for probabilistic logic gates are 42 followed by 213, whereas deterministic logic gates use 43 followed by 212 (other gates use other codons). The choice of start codons is arbitrary, while `42' was picked in homage to Douglas Adams's: ``The Hitchhikers Guide to the Galaxy"~\citep{adams2012complete}, where `42'  represents the ``Answer to the Ultimate Question of Life, the Universe, and Everything".
It is unwise to choose `0' or `1' as a start codon, as this would imply that large sections of the probability table would be interpreted as start signals, thus giving rise to overlapping genes that are hard to interpret. Similarly, it would have been unwise to create start codons consisting of three bytes, as new gates would be formed too rarely. 

When a start codon is found close to the end of the genome, the missing bytes are read from the beginning, making the genome effectively circular. If a start codon is found inside the coding region of another gate, another gate is made accordingly, leading to overlapping ``reading frames". This device, often found in nature to protect genetic information, is also useful in the evolution of MBs.
In principle, genomes can be organized into many chromosomes, as well as being diploid or polyploid, even including genders or other variations known to occur in nature. 

\subsection{Mutations to the Genome}
Since a MB is encoded by a genome, it is the genome that becomes transmitted to the next generation, and it is the genome that can experience mutations. After mutations are applied, the genome becomes translated into the brain again, and all changes manifest themselves as differences to the MB.

When a genome is copied, we allow for point mutations, insertions, and gene duplications. Each site has a chance to mutate at every replication event. In addition, there is typically a 20\% chance for a section of the genome to be deleted randomly. The size of that section is between 256 and 512 sites (about 300 sites are needed to encode for probabilistic logic gate for example, while a timer gate needs about 15 sites). Deletions can only occur if the genome is larger than 1000 bytes. Similarly, there is a 20\% chance for a section of 256 to 512 sites to be copied and randomly inserted anywhere in the genome (similar to a transposition event in genetics). This is not allowed if the genome is already above 20,000 sites. 

Both limits on insertions and deletions prevent genomes from disappearing or becoming intractably large. Note that the lower and upper limit of the genome size can be set by the experimenter. In principle, other kinds of mutations (including whole-genome duplications) are possible, and the hard-coded lower and upper limits of genome size are just practical values that can be changed arbitrarily. 

\section{Evolutionary Optimization}
MBs are typically used to control an agent or to act as a classifier or decision maker. In order to optimize the performance of each MB, the performance needs to be quantifiable so that MBs can be ranked in performance. Generally speaking, MBs that perform better will be given more offspring than poor performers. This essential dynamic in Darwinian evolution can be implemented in many different ways, the standard `Moran process' or `Tournament selection' are just two examples. Once a MB genome is selected for replication, it experiences the mutations described above. The quantity and quality of these changes and how they are applied (genetically or directly) is controlled by the experimenter. Typically, evolution needs thousands of generations for a population of 100 MBs to converge on a solution. The types of gates that are utilized plays an important role in shaping the adaptive trajectory. For example, we have experienced that for most behavioral tasks, deterministic MBs evolve faster than those that are made from stochastic elements.

\section{Analysis Methods}

\subsection{Visualizing a MB}
\begin{figure*}[b!ht]
\begin{center}
\includegraphics[width=6in]{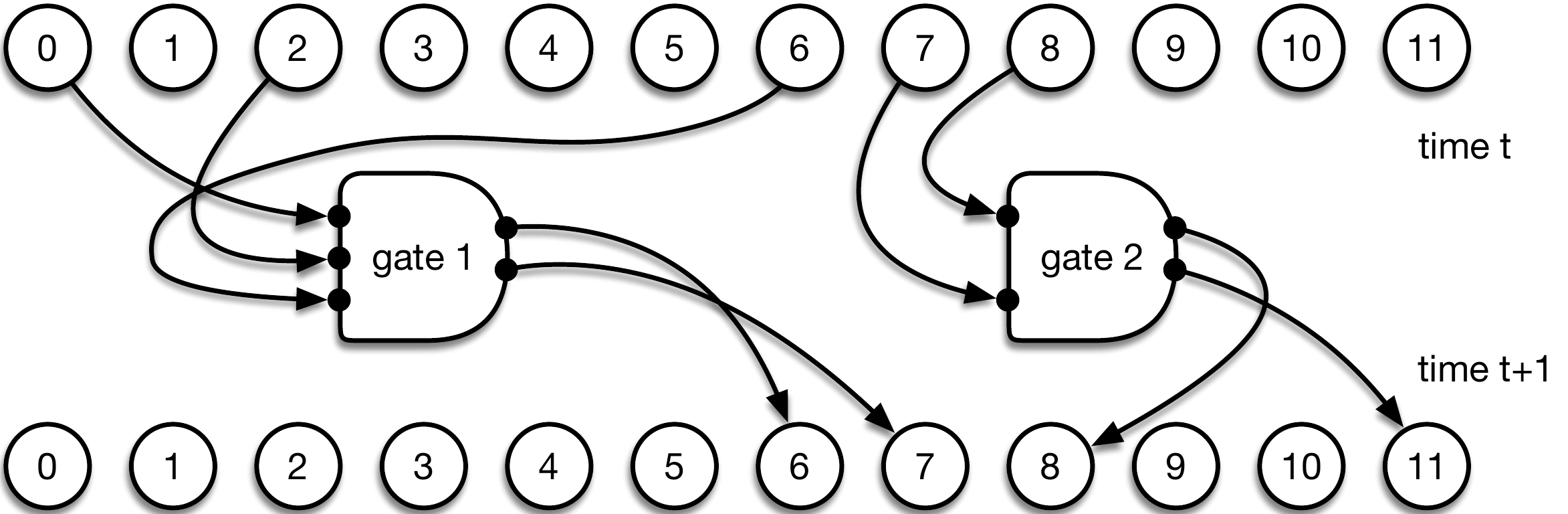}
\caption{Illustration of a MB with 12 states and two gates.}
\label{fig:connectivity}
\end{center}
\end{figure*}

We use several different ways to visualize a MB, and typically the idea is to give the user some information about the placement of gates and how they connect to nodes, or to show the user how nodes relay information to each other. The first way (see Figure \ref{fig:connectivity}), illustrates the state buffer at time point $t$ on top of the figure and the updated buffer at time point $t+1$ at the bottom. All gates are added as boxes, and if gate types matter, either colors or additional kinds of shapes can be used. The input connections for each gate are then drawn between the upper state buffer, while the output connections are drawn to the bottom state buffer. This symbolizes how information changes from update to update.

\begin{figure}[!htb]
\begin{center}
\includegraphics[width=2.0in]{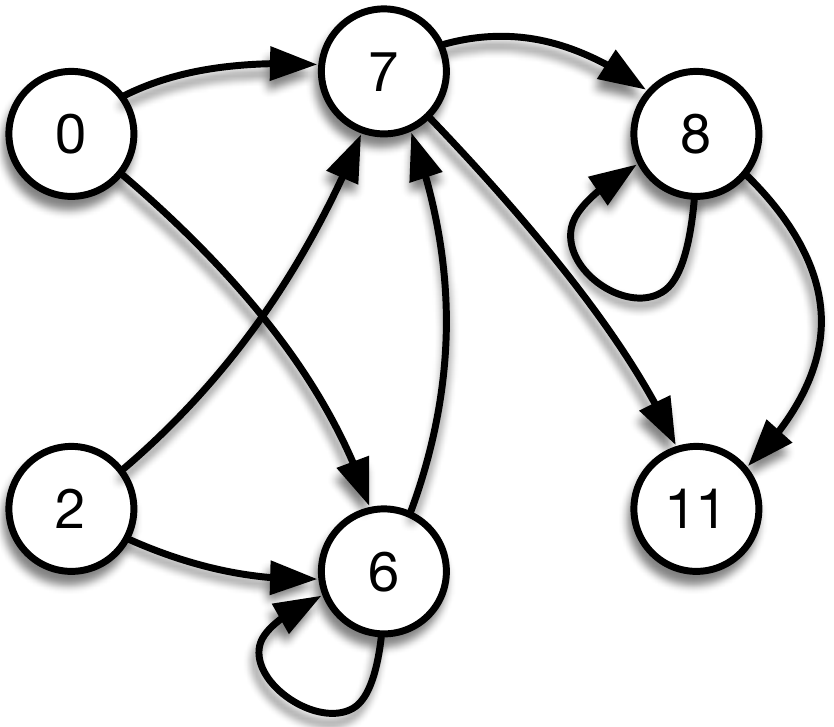}
\caption{Illustration of the same MB used in figure \ref{fig:connectivity}, but as a graph that shows only the nodes of the state buffer and how gates relay information between them.}
\label{fig:network}
\end{center}
\end{figure}

Alternatively, if the gates and their function or type is of lesser importance, the same MB can be represented as a graph, only showing the states and how they influence each other (see Figure \ref{fig:network}). This type of illustration also removes all nodes that do not have a connection and are therefore meaningless to the computation. In order to illustrate which nodes are sensors, and which ones are actuators, colors or shapes can be used.

\subsection{State-to-State Transition Diagrams}
\begin{figure*}[!htb]
\begin{center}
\includegraphics[width=6in]{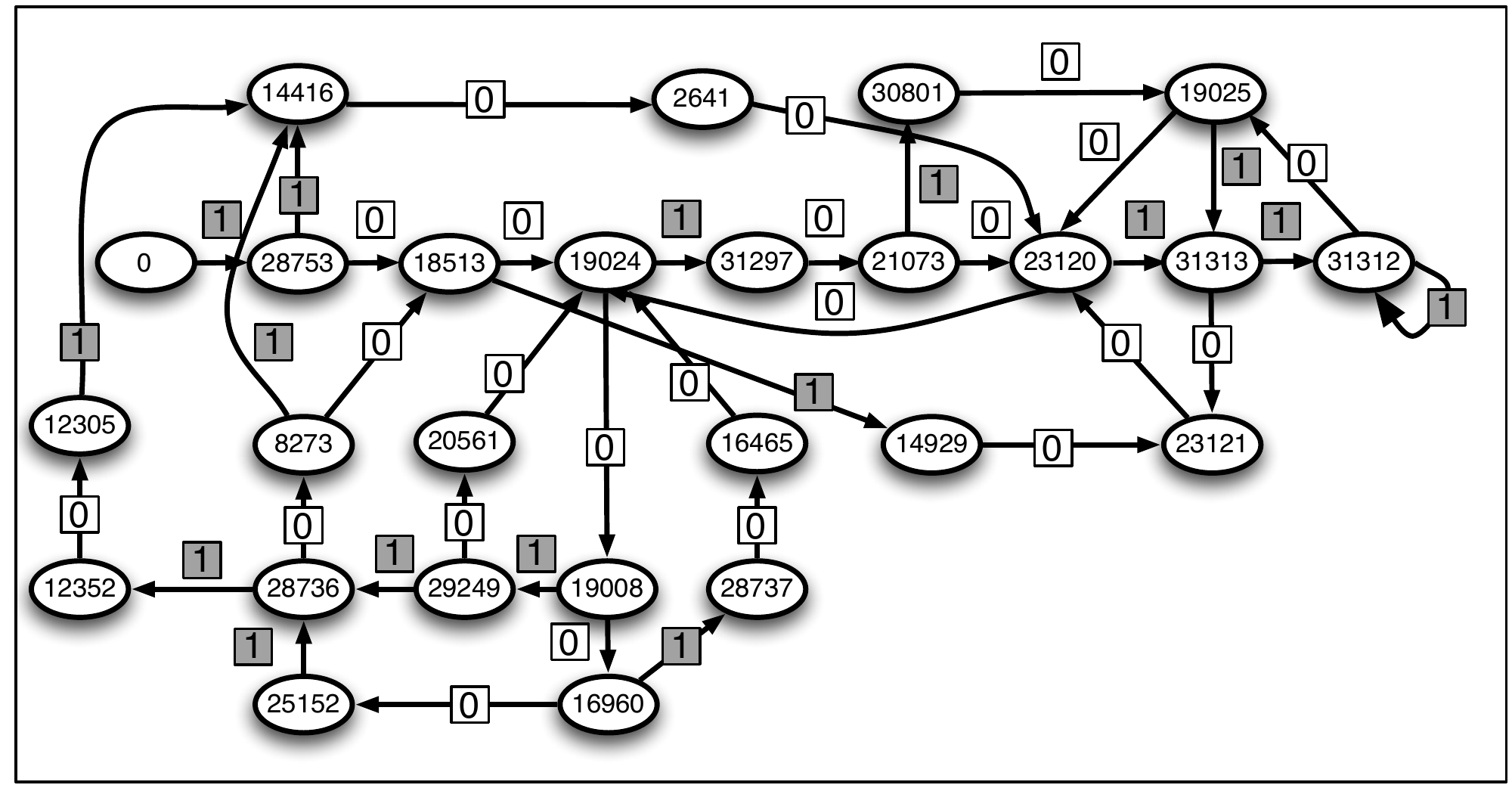}

\caption{State-to-state transition diagram for a 16-node brain that listens to a periodic tone (different lengths and periods), delivered to a single input. State labels are constructed as the decimal equivalent of the binary state vector of the functional brain (with inert nodes removed), not including the neuron that determines the transitions, whose state is indicated by boxes next to the arrows. This state-to-state diagram depicts only the transitions that are relevant to the Brain's fitness, which are the transitions shaped by evolution. A Brain is initialized in the quiescent state, which translates to the state `0'.}
\label{fig:s2s}
\end{center}
\end{figure*}

Over the lifetime, the state of a MB constantly changes. Recording the state of a Brain at every update allows the reconstruction of the state-to-state transition matrix that depicts the probability that a particular state is followed by a particular other state. In case continuous values are used as brain states a discretization step has to be performed first. In deterministic MBs, the state-to-state transitions are only affected by the sequence of inputs, and not by the probabilities of the computational components. Fig.~\ref{fig:s2s} shows an example state-to-state transition graph for a deterministic MB with a single auditory sensor (a brain that listens to a periodic tone, tasked to make decisions conditional on what it hears). In this simple case, the transition is only influenced by whether the tone is present (``1'') or absent ``0'', and each input 
usually moves the brain from one particular state to another (unless the brain does not pay attention to the tone). Often (as in Fig.~\ref{fig:s2s}) states are labelled by the decimal equivalent of the binary state vector. Thus, prior to experiencing any input, the Brain begins in state `0'. In the example in Fig.~\ref{fig:s2s}, hearing a `1' moves this Brain (with a state vector of 15 nodes) from state 0 to state 28,753, which stands for the activation pattern $110110011001101$. It is important to note that the state-to-state diagram implies that the Brain's future behavior is strongly influenced by what state it is in, and that therefore its computations are highly dependent on its past experience. In this example, a particular sequence of tones will ``put the brain into a particular state", which will determine how it will respond to future tones as it anticipates only a few different future tones from this state. This state-dependence of computations makes MBs qualitatively different from standard (non-recurrent) ANNs. While recurrent ANNs can also perform state-dependent computations, in these networks all nodes of one layer are connected to all other nodes of the next, which implies that the current state of an ANN is best described as a {\em conglomerate} (an average), which gives rise to the large basins of attraction that are precisely what makes these networks good at classification tasks. In MBs, on the other hand, connections are sparse and thus states are much more interdependent and conditional on each other. As a consequence, the state of a MB can be changed much more easily, which makes MB in our experience a better substrate for fine-grained behavior, as well as better suited for the integration of temporal information.

When Brains receive inputs from multiple sources, state-to-state transition diagrams focusing on a single source can reveal important dynamics, but either must keep the state of other inputs fixed, or else show different arrows for each combination of inputs.

\subsection{Brain Size and Complexity Estimates}
While for natural organisms we can weigh a brain or estimate the number of cells involved, we really cannot easily estimate a brain's complexity. It is not trivial to estimate the size or complexity of any particular MB either, because there are many different quantities that measure different aspects of the Brain's performance.  Is the number of gates analogous to the number of neurons or is it the size of the state buffer that determines the brain's performance?

In fact, this problem is no different than other quests to define functional complexity. We could use Brain ``size" as a proxy for complexity, but the size of the state buffer is predefined and thus limited (a limitation we already removed for some implementations). Also, the maximum size of the genome is limited and thus affects complexity as well. Lastly, because elements of the Brain do not cost anything, increasing the number of components may not reflect functionality. 

Regardless of these caveats, it is possible to count the number of gates, the number of hidden states used, the diameter of the causal graph (see Figure \ref{fig:network}), the number of connections, or the size of the genome, as a proxy to assess Brain complexity. In principle, it would be useful to define the equivalent of a Vapnik-Chervonenkis dimension for MBs to measure the capacity of the Brain to learn new functions, but it is not a priori clear that such a capacity is the right way to quantify MB performance because brains must do more than just store patterns: they have to know things so that they can behave appropriately.

\subsection{Brain Density Estimates}
In addition to Brain ``size", we might ask how interconnected these Brains are. The general idea is that a sparsely connected brain does not relay a lot of information between computational units, whereas a fully connected brain allows all computational units to have access to all information. It is not clear if this distinction has an actual impact on the computations performed. We define the density of a MB as the actual number of connections between states and the total possible number of connections between states. A connection in this context is the number of connections between nodes in the state buffer. Specifically, we count two nodes as connected if a gate relays information from one node to the other. Multiple connections between states are counted as one connection, otherwise an infinite amount of connections would be possible. Nodes that do not receive inputs from gates, and nodes that are not read from by gates are removed and do not count as having connections, since their state is either always 0 or does not contribute to the computation.

\subsection{Knockout Analysis}
A genetic knockout is a powerful tool to find causal functional relations within a Brain. The idea is to define a component's function by comparing the ``wild-type" behavior of the system to the behavior of the system after a component has been knocked out, just as in classical genetics. Knockouts can be performed on a MB on different levels. Sections of the genome can be removed, gates can be removed, or states can be silenced. While it is obvious how gates and parts of the genome can be knocked out, this is not simple for modifying states. A state cannot just be removed, since it would now leave gates with connections to nowhere. Therefore, the next plausible thing to do would be to set all states to $0$. Alternatively, the state could be set to $1$, or a random number for that matter. The loss of function would depend on the kind of knockout performed. 

\subsection{Neuro-correlate \texorpdfstring{$\Phi$}{Phi}}
The ability of a MB to integrate information can be measured using the neuro-correlate $\Phi$ \citep{Balduzzi2008, Oizumi2014}, which is a measure of how well a particular network integrates information (see below for references). Calculating the $\Phi$ of a network is a rather complex and time-consuming procedure, and requires the user to record the state-to-state transitions over the lifetime of the MB, and for later $\Phi$-measures \citep{Oizumi2014}, the connectivity of the MB. However, MBs were instrumental in validating this measure as a possible way to measure brain complexity~\citep{edlund2011integrated,joshi2013minimal,albantakis2014evolution}.

\subsection{Neuro-correlate \textit{R}}
Another possible way of estimating brain complexity is to quantify how much of the world is represented within the MB's internal states, meaning, how much the brain knows about the world without consulting any of its sensors. We have shown that a quantitative measure of representation, $R$, is highly predictive of the brain's fitness (and therefore function)~
\citep{marstaller2010measuring,marstaller2013evolution}).
It is important to note here that the work on representation showed that MBs do not evolve to use a single state to store a specific piece of information, but that instead information is smeared--or distributed--over many states, and that the concepts that are being represented within these groups of states are generally not simple concepts, but rather compound concepts. 

\subsection{Modularity}
The modularity of MBs can be assessed using the graph representation of the Brain (see Figure \ref{fig:network}). However, we found that this only gives meaningful results once Brains are larger than 100 nodes. Also note that such modularity measures would only quantify topological modularity, not functional modularity because to do the latter, nodes would need to be assigned to particular functions. When performing functional analysis and knockouts, we find that states contribute to many functions, which makes it impossible to use generic modularity measures in which nodes must be assigned to one function only. One way out of this dilemma is to use the modularity measure $Q_{H}$\citep{hintze2010modularity} which allows the user to measure modularity for systems whose components are in several modules at the same time.

\subsection{Mutational Robustness}
\begin{figure}[!htb]
\begin{center}
\includegraphics[width=3in]{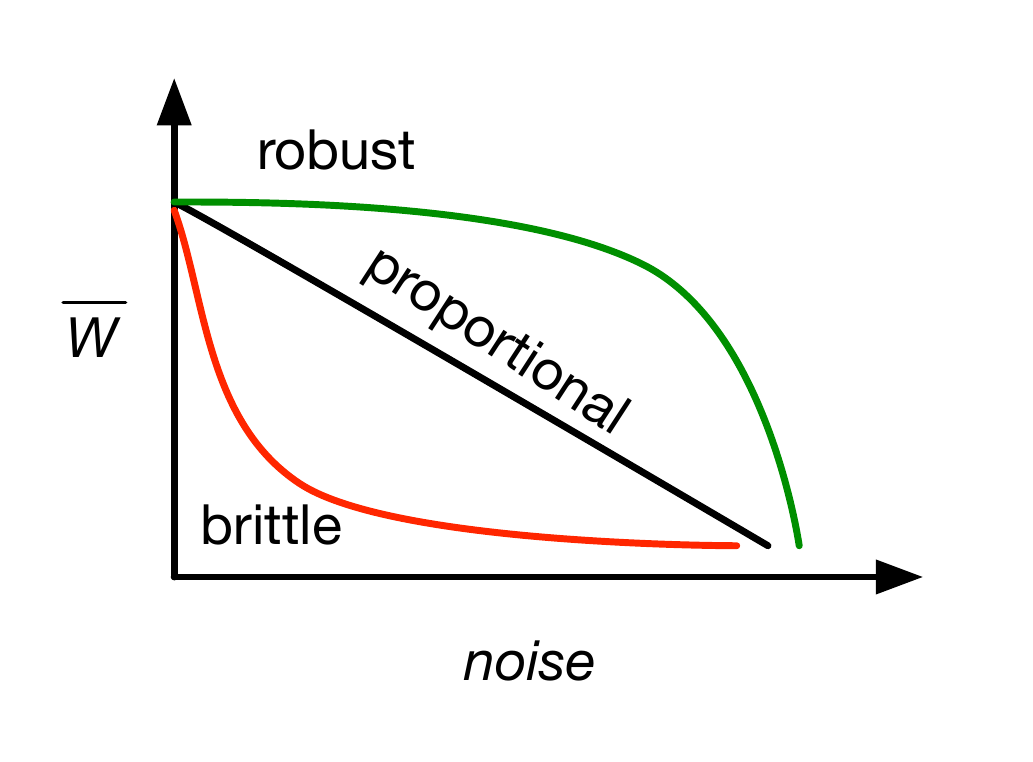}
\caption{Illustration of mutational robustness, the average fitness ($\bar{W}$) is shown as a function of the number of mutations or noise experienced (black). If this curve remains high (green) for more mutations it indicates that the system is very robust. A quick loss of functionality indicates a brittle system (red).}
\label{fig:robustness}
\end{center}
\end{figure}

Mutational robustness is a measure that seeks to assess if a MB can tolerate multiple mutations, or if it is very brittle. To this end--either exhaustively or by sampling--the fitness of a MB is measured when experiencing single, double, triple, and even more mutations at the same time. When plotting the mean fitness over the number of mutations, we can assess the robustness of the system (see Figure \ref{fig:robustness}). This also works for arbitrary sources of noise, not just mutations. In those cases we would ascertain the robustness of the MB with respect to that particular form of noise.

\subsection{Quantification of Behavior}
Beyond just visualizing the agent controlled by a MB as a movie, behavior can also be quantified differently. While we highly recommend that each experimenter observes the behavior of an agent visually if at all possible, we need to point out that such an assessment is subjective, not quantitative, and at best anecdotal. 

The simplest way of collecting behavioral data is to record the sequence of all actions taken, given the sequence of inputs received. The changes in frequencies of actions can be indicative. Also, which action is followed by which other action might be an interesting source of information. 

Beyond the sequence of actions, we have also deployed other measures to assess the behavior of agents, for example how much they swarm \citep{olson2013predator}, that is, act collectively as a group. Other measures such as distance travelled, food collected, or any other property that can be recorded might be useful. The important concept we try to relay here is that any type of quantifiable data together with a properly executed statistical analysis is better than any subjective observation and interpretation of behavior, much as in standard behavioral biology.

\section{History}
It is not clear from a search of the literature to what extent MBs were already defined in this particular way before, and whether they have been used or evolved earlier by others. The concept of a probabilistic finite state machine controlled by inputs is implicit in any hidden Markov model of a process. A standard example are the Hidden Markov models (HMMs) trained to recognize--or even synthesize--particular molecular sequences, and the name we chose for the technology refers directly to this heritage. The development of MBs owes a great deal to the insights of Jeff Hawkins (discussed for example in his book ``On Intelligence"~\citep{Hawkins2004}), who correctly views human brains as prediction machines that are deeply and fundamentally conditioned by their past experience. Their recurrent nature, as well as the idea to create a complex system with an extreme bandwidth that is still being made from individual computational units was inspired by the book ``I am a strange loop"~\citep{hofstadter2008strange} from Douglas R. Hofstadter.

The fact that many different computational implementations result in Turing complete machines implies that it is always possible to implement the functionality of system using a different technology. When a MB is a network of deterministic gates, it could in principle implement an IBM 8088 chip, and therefore simulate a universal computer. Whether evolution ever would optimize MBs towards such an architecture is, however, extremely unlikely.

Ultimately, we should think of a MB as a finite state machine, whose current state is defined by the state buffer, which becomes updated according to the inputs and the current states (much as in Bayesian inference) but with access to internal models for shaping the priors. If the computational components are deterministic the MB would resemble a deterministic finite state machine, while a MB made from (or at least containing some) probabilistic computational components would resemble a probabilistic finite state machine.

It is, as a consequence, difficult to find a proper classification of MBs. Are they a strange implementation of a finite state machine, or a weird implementation of a Hidden Markov process? Also, where does the definition of a MB start and end? Does it necessarily include a genetic encoding that is very flexible and can be varied, or could it even be ignored in the case of direct encoding? Regardless of an actual answer, MBs and a method to evolve them have now been implemented multiple times already, independently, and for various purposes. Here we will give a brief historic overview of those implementations.

\subsection{Integrated Information}
The first implementation was written by Jeffrey Edlund, and this installment used a stack of genes without non-coding regions, instead of a full genome that uses start codons and thus could have significant stretches of non-functional code. This implementation was quickly changed to the genetic encoding in the same project. Without having performed a quantitative comparison of both systems, the consensus was that the genetic encoding (using start codons) evolved faster, and to higher fitness, on average. This implementation also used only one type of start codon, and only probabilistic logic gates. In 2011 we published the first article using this implementation of MBs \citep{edlund2011integrated}.

\subsection{Representations}
The original work on representation~\citep{marstaller2010measuring,marstaller2013evolution} has been expanded and now includes multiple different start codons that define gates both deterministic amd probabilistic logic gates, as well as a precursor of the artificial neuronal network gates. This framework was written by Arend Hintze, and has been expanded and adapted several times and for several other experiments \citep{hintze2014evolution,albantakis2014evolution,albantakis2015,kvam2015computational}. Adding threshold and timer gates enabled the evolution of social hierarchies in digital agents~\citep{hintze2014evolution}.

\subsection{Swarming}
Allowing for the evolution of swarms as well as predator prey dynamic required the expansion of the framework mentioned above. All work that involves swarms of agents was performed using the extension written by Randal Olson made~\citep{olson2013predator,olson2013critical,haley2014exploring,olson2015elucidating,haley2015evolving,hintze2016orthogonally,olson2015exploring,tehrani2016flies,olson2016evolution,olson2016exploring,olson2012bottom}. This work also resulted in a Python implementation of Brains that can use deterministic and probabilistic logic gates \url{https://github.com/rhiever/MarkovNetwork}.

\subsection{Learning}
Studying the evolution of learning further expanded the framework. This work added feedback gates to study agents that were able to adapt to an environment within their lifetime, and was performed by Leigh Sheneman and Arend Hintze \citep{sheneman2017machine}. 

\subsection{EALib}
MBs were also added by David Knoester to EALib (EALib; \url{https://github.com/dknoester/ealib}), which is a software library to develop a wide range of different evolutionary computational experiments. This framework was used in several publications since then \citep{chapman2013evolution,chapman2017evolution,goldsby13Dirty,goldsby2014effect,goldsby2017increasing,knoester2012constructing,knoester2011neuroevolution}. This implementation allowed for much larger brains (thousands of states) and used a somewhat different encoding of gates.

\subsection{MABE}
Two key insights about software development that have emerged within the last decade are the importance of platform independence, and the importance of code ``re-cycling", that is, to enable an environment where code can easily be re-purposed, akin to ``horizontal transfer" of genetic code in biological organisms. Creating a platform that is specific to MBs runs the risk of producing results that are germane to their particular implementation, and thus question the generality of the results obtained. Therefore, ideally experiments should be repeated using other types of computational brains, and a modern platform should provide a level of modularity that makes this easy. Secondly, code that was developed for one experiment should be used in other experiments (possibly suitably modified), either as a control or as an improvement. This only works if the code is designed to be modular, and to explicitly allow for this kind of exchange without it breaking earlier constraints. 

MABE~\citep{bohm2017mabe}, the \ul{M}odular \ul{A}gent \ul{B}ased \ul{E}volver framework, is based on these two principles: substrate independence and modular design. MABE supports all prior experiments, and contains some of them as examples. The platform is designed to support a vast variety of future experiments, and most importantly it is designed in such a way that contributions from one experimenter can be used by others. This mix and match (or plug-and-play) philosophy already allowed for a set of other experiments~ \citep{hintzeBartlettDyer2015,schossau2015information,schossau2017role}, and we suggest that future experiments should be done using MABE.

\section{Acknowledgements}
This work was supported in part by NSF's BEACON Center for the Study of Evolution in Action, under Contract No. DBI-0939454 and  by the Paul G. Allen Family Foundation. We are grateful to Christof Koch and Giulio Tononi for support, discussions, and crucial insights. We thank Fred Dyer, Frank Bartlett, Lars Marstaller, Devin McAuley, Heather Eisthen, Nikhil Joshi, Samuel D. Chapman, Tim Pleskac, and Ralph Hertwig for attempting to keep us honest with respect to neuroscience, as well as the intelligence and behavior of real animals (including people).

\footnotesize

\end{document}